\documentclass[journal]{IEEEtran}

\ifCLASSINFOpdf
\else
\fi

\usepackage{times}
\usepackage{epsfig}
\usepackage{graphicx}
\usepackage{amsmath}
\usepackage{amssymb}
\usepackage{algorithm}
\usepackage{algorithmic}
\usepackage{subfiles}
\usepackage{multirow}
\usepackage{subcaption}
\usepackage{tablefootnote}
\newcommand{\etal}{\textit{et al}.}

\newcommand{\eg}{\textit{e}.\textit{g}.}
\usepackage{balance}
\usepackage{cite}

\DeclareMathOperator*{\argmin}{argmin}

\usepackage[colorlinks,linkcolor=red]{hyperref}
\usepackage{makecell}
\usepackage[table ]{ xcolor}
\definecolor{mygray}{gray}{.9}
\definecolor{white}{gray}{1}
\usepackage{color}
\usepackage{enumitem}
\usepackage{url}
\usepackage[square,sort&compress,numbers]{natbib}

\begin{document}
%
\title{P2P-Loc: Point to Point Tiny Person Localization}
%
%
%

\author{Xuehui Yu, Di Wu, Qixiang Ye, Jianbin Jiao and Zhenjun Han 
\thanks{Corresponding authors: Zhenjun Han.} 
\thanks{X. Yu, D. Wu, Q. Ye, J. Jiao and Z. Han are with the School of Electronic, Electrical and Communication Engineering, University of Chinese Academic of Sciences (UCAS), Beijing, 101408 China. E-mail: \{yuxuehui17, wudi20\}@mails.ucas.ac.cn, \{qxye, jiaojb, hanzhj\}@ucas.ac.cn.}
}

%
%

\maketitle
\begin{abstract}
Bounding-box annotation form has been the most frequently used method for visual object localization tasks.
However, bounding-box annotation relies on a large amount of precisely annotating bounding boxes, and it is expensive and laborious. It is impossible to be employed in practical scenarios and even redundant for some applications (such as tiny person localization) that the size would not matter.
Therefore, we propose a novel point-based framework for the person localization task by annotating each person as a coarse point (CoarsePoint) instead of an accurate bounding box that can be any point within the object extent. Then, the network predicts the person's location as a 2D coordinate in the image.
Although this greatly simplifies the data annotation pipeline, the CoarsePoint annotation inevitably decreases label reliability (label uncertainty) and causes network confusion during training. As a result, we propose a point self-refinement approach that iteratively updates point annotations in a self-paced way.
The proposed refinement system alleviates the label uncertainty and progressively improves localization performance.
Experimental results show that our approach has achieved comparable object localization performance while saving up to 80$\%$ of annotation cost. 

\end{abstract}


\IEEEpeerreviewmaketitle

\section{Introduction}
\IEEEPARstart{O}{bject} localization is essential in the computer vision community with various systems, including visual surveillance, driving assistance, and mobile robotics  \cite{enzweiler2008monocular,dollar2011pedestrian,geiger2012we,zhang2017citypersons,mao2017can,DBLP:journals/tits/HavyarimanaXSWB20,DBLP:journals/tits/YinWDTHX20,DBLP:journals/tits/ChoiK20,DBLP:journals/tits/ThangarajahW20,zhang2017towards}. Object localization has achieved unprecedented progress with the rise of deep learning and large-scale bounding-box or precision point annotations.

\begin{figure}
\begin{center}
    \begin{tabular}{ccc}
    \includegraphics[width=8.5cm]{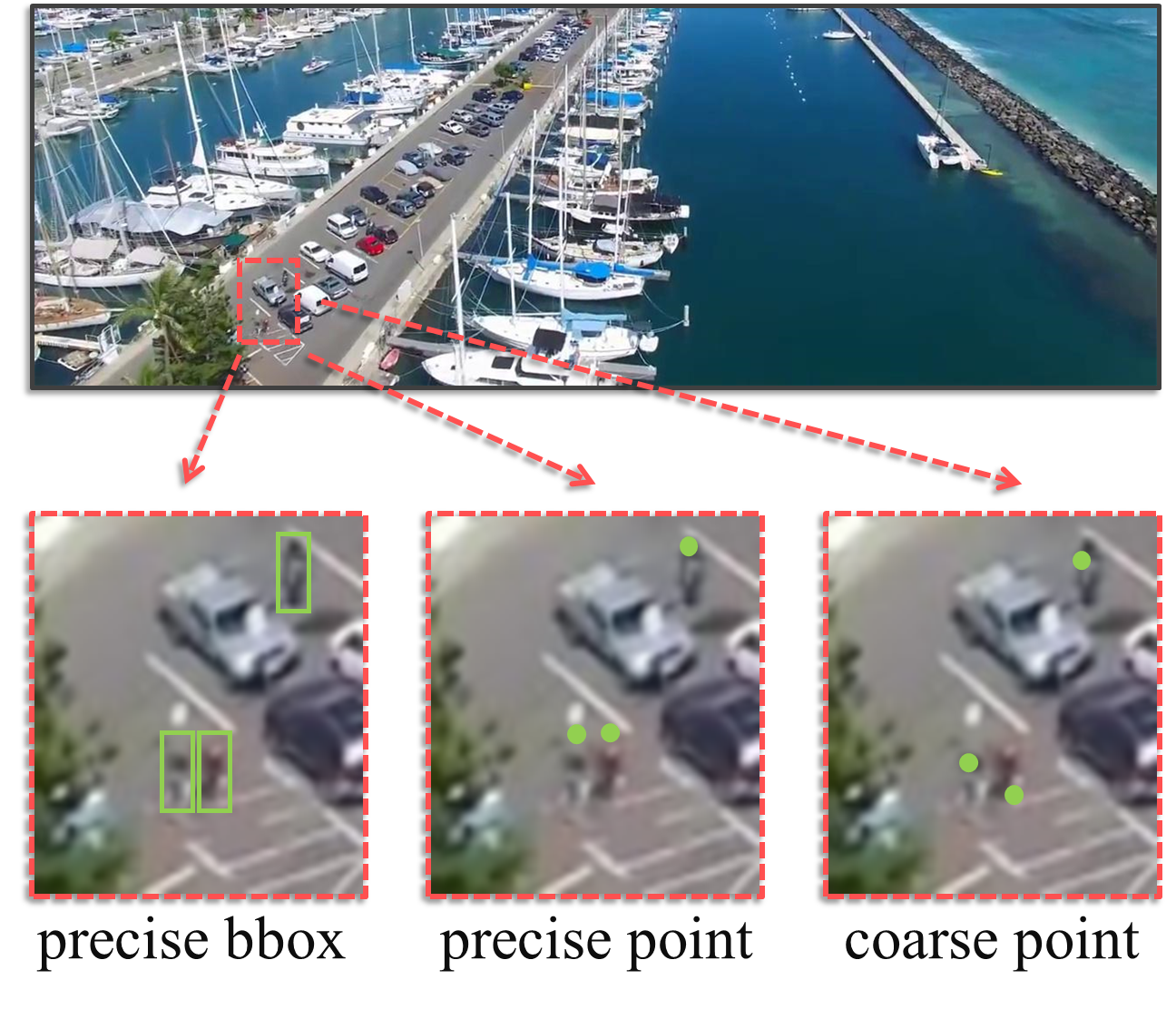}
    \end{tabular}
   \caption{Comparison of different annotation forms.
   A precise point may be a person's head or the center of a bounding box. In some scenarios, we only focus on the coarse location of the object where the three annotation forms have the same effect.  
   }
\label{fig:anno_form}
\end{center}
\end{figure}

Precise object location usually requires precise annotation, as shown in Fig.~\ref{fig:anno_form}.
To pursue high-precision positioning, a general object detection presents each target as a tight bounding rectangle. For example, the pose estimation models the human body as seventeen key points in fixed positions. 
In scenarios where targets are far away or of low resolution~\cite{Yu2020ScaleMF,DBLP:conf/cvpr/XiaBDZBLDPZ18,DBLP:journals/tits/HanWYG20}, it is time-consuming and even infeasible to implement precise annotation due to the small object size and low signal-to-noise ratio. 

Therefore, we creatively propose a new framework that can localize objects with easily acquired supervision signals.
In the low-resolution scenarios, the framework is intended to locate the object's position with needless bounding boxes. 

\begin{figure*}
\begin{center}
    \begin{tabular}{ccc}
    \includegraphics[width=13cm]{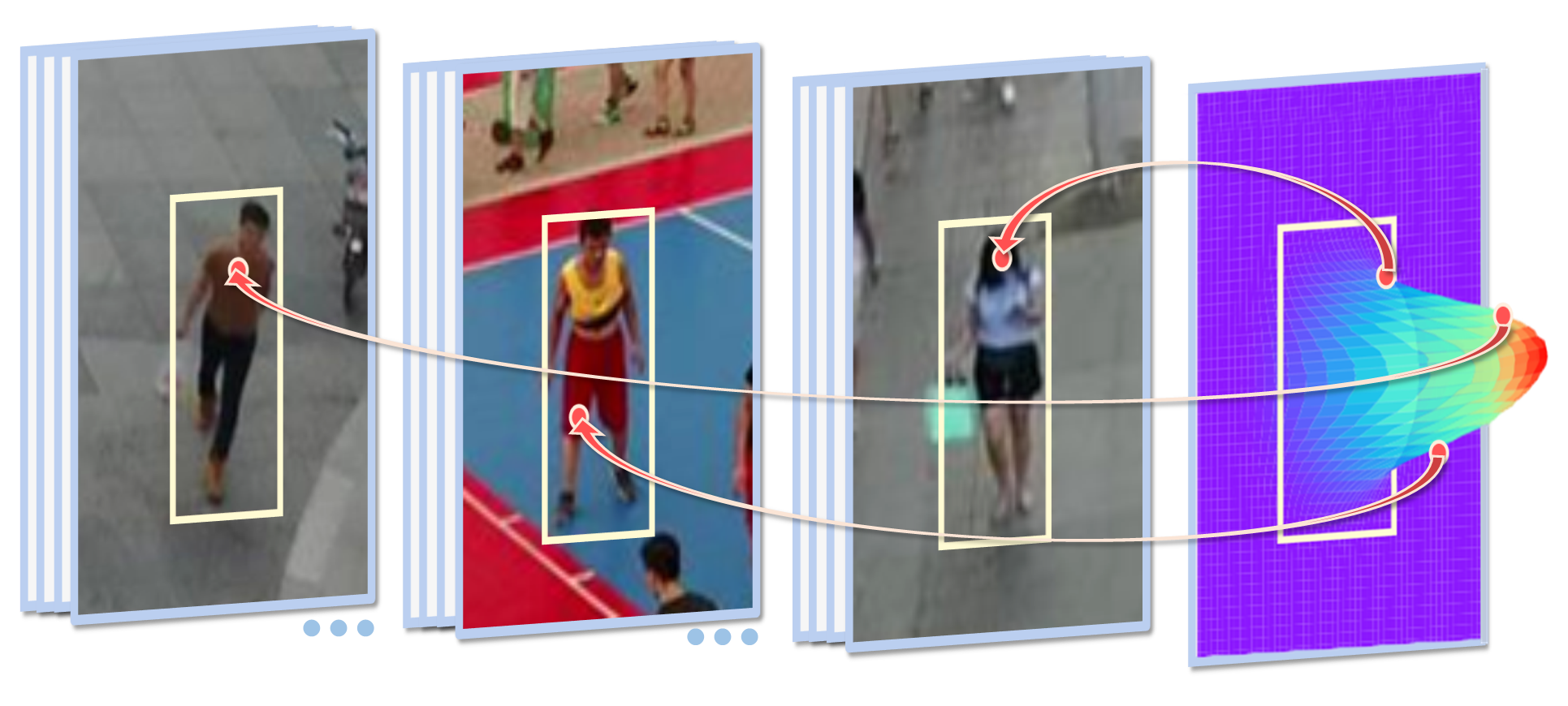}
    \end{tabular}

   \caption{Challenge of the CoarsePoint annotation. The same part of different instances can be annotated as positive or negative. We take Rectified Gaussian Distribution as an example. Located on the far right is a schematic diagram of the labeled points on the position distribution of the entire data set. More details are provided in Sec.~\ref{5.2}.}
\label{fig:uncertainty}
\end{center}
\end{figure*}

In this paper, we propose a novel computer vision task to achieve object localization with coarsely annotated point supervisions, referred to as CoarsePoint.
Because any point in the object area can serve as a CoarsePoint, the required human effort is greatly reduced for data annotation.  
However,  training with inaccurate supervision information will deteriorate the model due to the label uncertainty, such as the significant variance of annotated instance appearance and features uncertainties.
For example, different parts of a human body (\emph{e.g}., head, torso, or foot) could be labeled as positive instances in the training set, and other parts will be set as negative examples, shown in Fig.~\ref{fig:uncertainty}. 
The heads of different instances may be labeled as positive or negative samples. 
The label uncertainty brings two negative impacts: On the one hand, the same parts in different instances could be annotated as either positives or negatives, which misleads the localization model toward undesirable tendencies; On the other hand, different object parts may be labeled as positives for objects of the same kind, which will propel the network to predict different parts of the same object, aggregating the risk of false positives.

To tackle the negative impact of CoarsePoint,  we propose a simple-yet-effective strategy point self-refinement, aiming at narrowing the performance gap between the coarse point supervision and the precision point supervision. 

The idea behind self-refinement is that when coupled with localization model learning (\eg, CNN training), the coarsely annotated points can be iteratively promoted in a self-paced fashion.
Self-refinement investigates the statistical characteristics of annotations and semantic features, pursuing a gradual decrease of label uncertainty and better convergence of model training. 

The contributions of this study include that:

\begin{itemize}

    \item We propose a CoarsePoint vision task, and under the relaxed point supervision, we set the first solid baseline for object localization;

    \item  
    We propose the self-refinement approach to promote coarse points, implement the statistical stability of supervision signals, and improve the localization model in a self-paced fashion;
    \item CoarsePoint achieves competitive experimental results with precise bounding-box annotation-based methods while saving up to 80$\%$ of annotation cost.
\end{itemize}


 

\section{Related Work}
Objects can be modeled in different fashions based on semantic and geometric characteristics, \eg, bounding-box, point, and so on.

To better explain our proposed task, we review related works from different supervision and evaluation fashions, respectively.

\subsection{Object as Bounding-box} 
Different from the CoarsePoint, the bounding-box-based modeling represents the position and scale information.
It can be categorized into fully supervised and weakly supervised methods based on the different supervisions.

\noindent\textbf{Fully Supervised Methods.} General object detection models object as a bounding box. As instance-level annotation, a bounding box is created to render the center and scale of the object. 
Under this setting, more and more fully supervised detectors\cite{FasterRCNN2015,FPN2017,CenterNet2019,DBLP:journals/tits/YeZZL21,DBLP:journals/tits/HassaballahKMM21,DBLP:journals/tits/YangZWXDY21,DBLP:journals/tits/CamaraBCNAWRDF21,DBLP:journals/tits/CamaraBCWNAWRDM21,DBLP:journals/tits/BaekHK20} are expected to represent the object accurately. However, such supervision form would consume huge workforce physical resources, and it takes much time and heavy workload to annotate the object at the instance level.

\noindent\textbf{Weakly Supervised Methods.} To lower the annotation cost and make full use of massive network data, the weakly supervised object detection (WSOD) \cite{DBLP:conf/cvpr/BilenPT15,DBLP:conf/cvpr/BilenV16,DBLP:conf/icml/SongGJMHD14,DBLP:conf/iccv/SivaX11,DBLP:journals/tip/WangHRZM15} trains a detector when only image-level labels are available, which will cause WSOD to focus only on local areas and lack the ability to distinguish instances due to the lack of instance-level constraints. 
Different from WSOD, weakly supervised object localization utilizes the activation map from the last convolution layer to generate semantic-aware localization maps for object bounding-box estimation~\cite{zhou2016learning}.

The evaluation of the above tasks is all based on the bounding box. Average precision (AP) and the Correct Localization (CorLoc) \cite{DBLP:journals/ijcv/DeselaersAF12} are used to evaluate detector performance.
Unlike the methods mentioned above, CoarsePoint has no downstream task and does not focus on the object geometry. We only focus on the location points and evaluate the performance according to the position of the predicted point, avoiding bounding box annotation. 

\subsection{Object as Point} 
For some vision tasks, it is not necessary to model an object in the form of a bounding box. Researchers began to pay attention to the annotation form of points that serve to represent part of or the entire object. 

\noindent\textbf{Human Keypoint Detection.} 
Human keypoint detection, also known as human posture estimation, aims to locate the position of joint human points accurately. 
The COCO dataset \cite{lin2014microsoft} contains over 200, 000 images and 250, 000 person instances labeled with 17 key points. 
During inference, the pose detector predicts the position of every visible keypoint. Similar to COCO, the Human3.6M \cite{DBLP:journals/pami/IonescuPOS14} is one of the most extensive 3D human pose benchmarks.
With this benchmark, the detector needs to predict the position of each keypoint in the 3D coordinate system. 
Three-point annotation form datasets used in \cite{DBLP:conf/cvpr/RiberaGCD19} are intended to locate the object accurately without considering other redundant information or reduce annotation burden by using the form of the center point~\cite{DBLP:conf/cvpr/PapadopoulosUKF17}.

\noindent\textbf{Object Localization.} A new task \cite{DBLP:conf/cvpr/RiberaGCD19} has been proposed to estimate object locations without annotated bounding boxes. In object localization, a true positive is counted when the distance between the estimated location and a ground truth point is less than \emph{d}, and vice versa. Then the evaluation results of precision and recall can be calculated. In addition, several evaluation metrics are adopted to indicate if the number of location points is incorrect, such as Mean Absolute Error (MAE), Root Mean Squared Error(RMSE), and Mean Absolute Percent Error (MAPE)~\cite{choe2020evaluating}.

Different from the above tasks, CoarsePoint does not require precise point annotations, which significantly reduces the annotation time. Meanwhile, the expectation of Coarse only aims at locating objects without extensive human effort, which does not require an extremely accurate localization effect like the above tasks. Therefore, the object is considered to be successfully located if the point falls into the corresponding bounding box. 

\noindent\subsection{Object as Counting} 

Crowd counting focuses on the number of people in the current scene rather than the position of human targets. In crowd counting, an accurate head annotation is utilized as point supervision. Generated by head annotation, a crowd density map is chosen as the optimization objective of the network. More importantly, On the contrary, CoarsePoint only focuses on the coarse position of the human body. The precision requirement of the annotation is relatively low.

\section{Methodology}\label{sec3}
Considering the statistical stability of the semantic center, we use the semantic center of the object to replace the coarse point annotation during the self-refinement process. First, on the training set, an estimator is trained under the coarse point supervision. Then, the estimator is used to predict the location of semantic statistic points (SSPs) on each image of the training set. Finally, to reduce the uncertainty of coarse points, the center of the SSPs' location of each object is chosen as the refined point annotation. To simplify the description, the following formulation in this paper is only for localizing one category object. In the case of multiple categories, object localization is treated as multiple tasks of one-category object localization.

\begin{figure*}
\begin{center}
    \begin{tabular}{ccc}
    \includegraphics[width=18cm]{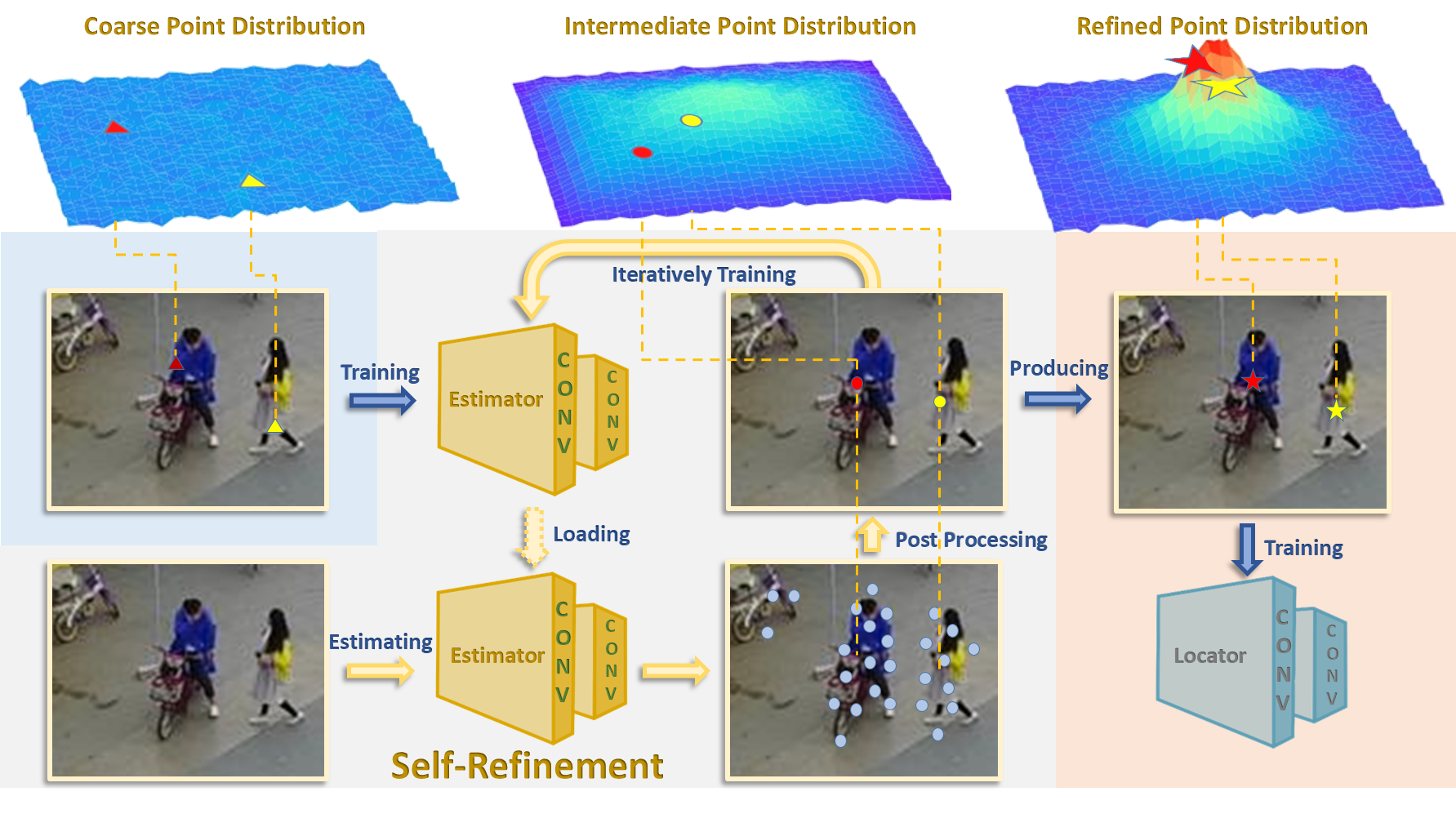}
    \end{tabular}

\vspace{-2mm}
   \caption{The pipeline of the self-refinement algorithm. 
   An estimator is trained under the coarse points supervision. The estimator is then used to find all SSPs in each image of the training set. Finally, the center of the SSPs of each object is chosen as the refined point annotation.
   After the first iteration, the refined annotation is the input annotation of the next iteration.
   A point is initially modeled as uniform distribution. With the iteration of the estimator, the point distribution gradually converges to the center of SSPs.
    }
\label{fig:framework}
\end{center}
\end{figure*}

\begin{algorithm}
\caption{Self-refinement Iteration}
\label{alg:URI}
\textbf{Input:} Training images $I$ \\
\textbf{Input:} Annotated Points $A$ \\
\textbf{Output:} Refined Points $A'$\\

\begin{algorithmic}[1]
\STATE $k = 0$, $A^0 = A$
\REPEAT
\STATE Train an estimator $E^k$ on $I$ with $A^k$ as label, obtain $E^k(\omega;\theta^*)$, Eq \eqref{estimator loss} \eqref{estimator train}
\FOR{each image $I_i$ in $I$}
    \STATE Inference on $I_i$ with $E^k(\omega;\theta^*)$, obtain $\hat{A_i^k}$, Eq \eqref{Eq:estimate point}
    \STATE Assign points in $\hat{A_i^k}$ to objects, obtain $G_{ij}$, Eq \eqref{Eq:assigning}
    
    \FOR{each points group $G_{ij}$}
        \STATE Merge $G_{ij}$ to obtain $A_{ij}^{k+1}$, Eq \eqref{Eq:merge}
    \ENDFOR
\ENDFOR
\STATE $k = k + 1$
\UNTIL{$k == MAX\_ITER\ or\ A^{k}== A^{k-1}$}
\STATE $A' = A^k$
\end{algorithmic}
\end{algorithm}

\subsection{Point Estimation} 
%
\noindent\textbf{Formulation of Semantic Parts.} The image set for training is $I=\{I_1, I_2, ...I_N\}$, coarse points sets are defined as $A=\{A_1, A_2, ...A_N\}$, in which $N$ is the number of images. $A_{i}=\{a_{i1}, a_{i2}, ...a_{iM_i}\}$.
$a_{ij}=(x, y)$ is the 2d coordinate of the annotated coarse point of the $j$-th object, and $M_i$ is the number of objects in the image $I_i$. The set of all points in $I$ is $\omega$ that is defined as $\Omega=\mathop{\cup}\limits_{i=1, 2..N}\{a|a\ is\ the\ point\ in\ I_i\}=\{\omega_1, \omega_2, ..\omega_h, ..\}$. Whether $\omega_h$ is annotated or not is denoted by $l_h$; $l_h=1$ means $\omega_h$ is annotated, whereas $l_h=0$ means it is not. 
Accordingly, the annotation set can be denoted as $L=\{l_1, l_2, ..l_h, ..\}$.

From a semantic perspective, objects can be divided into several semantic parts (\eg, a human object can be divided into head, hand, leg, \etal). $T$ donates the number of divided semantic parts, and $S_t$ $(t=1, 2...,T)$ represents the set of points within the $t$-th part of all these category objects in $I$. $S_0$ is
the set of points that are not in any part of the category object in $I$. All these point sets constitute $S=\{S_0, S_1, S_2, ..., S_T\}$, and $\mathop{\cup}\limits_{0\le t\le T}S_t = \Omega$ is the set of all points in $I$.

\noindent\textbf{Objective Function.}
The set of annotated points in $S_t$ is represented as $S_t^{+}$, and the annotated frequency of the semantic part $S_t$ is defined as $Q(S_t)$.

\begin{equation}
\begin{aligned}
&S_t^{+} = \{\omega_h |l_h=1, \omega_h \in S_t\}, \\
&Q(S_t)=P\{l_h=1|\omega_h\in S_t\}=\frac{|S_t^+|}{|S_t|}, \\
\end{aligned}
\end{equation}
where $|S_t^+|$, $|S_t|$ indicates the number of elements in $S_t^+$, $|S_t|$.

The objective function for training the estimator $E(\omega;\theta)$ under the supervision of $L$ is defined with Eq \eqref{estimator loss}:
\begin{equation}
\begin{aligned}
loss &(\theta;\Omega, L)= \frac{1}{|\Omega|}\mathop{\sum}\limits_{1\le h\le |\Omega|} FL(E(\omega_h;\theta), l_h)\\
& = \mathop{\sum}\limits_{S_t \in S} \frac{|S_t|}{|\Omega|} FL(E(\omega_h;\theta), Q(S_t)),
\label{estimator loss}
\end{aligned}
\end{equation}
where $FL$ is the focal loss, and the derivation process details of Eq \eqref{estimator loss} can be reached as follows with Eq \eqref{tmp_1} \eqref{appendix_1} \eqref{estimator train}:
  
\begin{equation}
\label{tmp_1}
\begin{aligned}
  SL_t^{-}=\{(f_h, s_h, l_h)| l_h=0, (f_h, sh, lh) \in SL_t\},
\end{aligned}
\end{equation}

\begin{equation}
\label{appendix_1}
\begin{aligned}
loss &(\theta;SL) = \frac{1}{|\Omega|}\mathop{\sum}\limits_{(f_h, s_h, l_h)\in SL} CE(E(f_h;\theta), l_h)\\
& = -\frac{1}{|\Omega|}\mathop{\sum}\limits_{(f_h, s_h, l_h)\in SL} [l_h*log(E(f_h;\theta)) +\\
& \ \ \ \ \ \ \ \ \ \ \ \ \ \ \ \ \ \ \ \ \ \ \ \ \ \ \ (1-l_h)* log(1-E(f_h;\theta))]\\
& = -\frac{1}{|\Omega|} \mathop{\sum}\limits_{S_t \in S} [\mathop{\sum}\limits_{(f_h, s_h, l_h)\in SL_t^+} log(E(f_h;\theta)) + \\
& \ \ \ \ \ \ \ \ \ \ \ \ \ \ \ \ \ \ \ \mathop{\sum}\limits_{(f_h, s_h, l_h)\in SL_t^-} log(1-E^{k}(f_h;\theta)))]\\
& = -\mathop{\sum}\limits_{S_t \in S} \frac{|S_t|}{|\Omega|} [Q(S_t)log(E(f_h;\theta)) + \\ & \ \ \ \ \ \ \ \ \ \ \ \ \ \ \ \ \ \ \ \ \ \ \
(1-Q(S_t)) log(1-E^{k}(f_h;\theta))]\\
& = \mathop{\sum}\limits_{S_t \in S} \frac{|SL_t|}{|\Omega|} CE(E(f_h;\theta), Q(S_t)),
\end{aligned}
\end{equation}

\begin{equation}
\begin{aligned}
\theta^* = \mathop{argmin}\limits_{\theta}\ loss(\theta;\Omega, L).
\label{estimator train}
\end{aligned}
\end{equation}
\noindent\textbf{Statistic Semantic Points.} For a point $\omega_h \in S_t$, the learning objective is to minimize the distance between the estimator $E(\omega_h;\theta)$ and the probability $Q(S_t)$ (instead of hard supervision, 1 or 0). To achieve a high score predicted by $E(a;\theta^*)$, there are two constraints:
(1) High $Q(S_t)$;  
(2) $S_t$ is semantically distinguishable from points of other categories and background ($S_0$). Such as SSPs.

In the $k$-th iteration, with images $I_i$ and the estimator $E^k(\omega; \theta)$ trained in $I$ under the supervision of $A^{k}$, the estimated SSPs can be obtained as Eq \eqref{Eq:estimate point}:


\begin{equation}
\hat{A_i^k} = \{\omega_h| E^k(\omega_h;\theta^*) > \delta, \omega_h \in \Omega\},
\label{Eq:estimate point}
\end{equation}
in which $\delta$ is a threshold value ranging from 0 to 1. In the experiment, $\delta$ is set to 0.2.

\subsection{Point Refinement}

The estimated semantic points $\hat{A^k_i}$ may come from different objects in the image $I_i$. They are assigned to the $j$-th object in $I_i$, noted as $G_{ij}$ in Eq \eqref{Eq:assigning}:
\begin{equation}
G_{ij} = \{a | a \in \hat{A_i}^{k}, j={\displaystyle \argmin_{1\le j' \le M_i}}{\ dis(a, A_{ij'})}\},
\label{Eq:assigning}
\end{equation}
in which $dis$ is a distance function to measure the Euclidean distance between two points.

\begin{figure}
\begin{center}
    \begin{tabular}{ccc}
    \includegraphics[width=9.5cm]{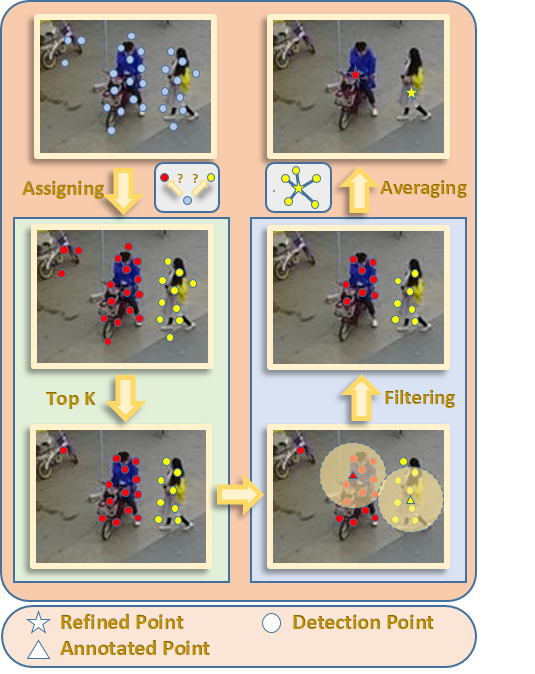}
    \end{tabular}
    \vspace{-2mm}
   \caption{The post-processing framework. The clustering method is used to assign which object the currently predicted point belongs to. To obtain a more precise label, we filter the top $K$ prediction by the distance threshold. The new label is obtained by the score weighting to merge the same kind of points.}
\label{fig:post processing}
\end{center}
\end{figure}

As Fig.~\ref{fig:post processing} shows, the top $K$ points in $G_{ij}$ ranked by the predicted score during the merging process will be kept. 
To obtain the final refined point annotations of the $j$-th object in image $I_i$, we calculate the average of the remaining points in $G'_{ij}$ with the score as its weight.
\begin{equation}
A_{ij}^{k+1}=\left\{
\begin{aligned}
\sum_{a \in G'_{ij}} \frac{score_a}{\sum_{a' \in G'_{ij}}{score_{a'}}} a & \ \ \ , G'_{ij} \neq \emptyset \\
A_{ij}^{k} & \ \ \ , G'_{ij} = \emptyset,
\end{aligned}
\right.
\label{Eq:merge}
\end{equation}
in which $G'_{ij}=\{a |a\in G_{ij},\ dis(a, A_{ij}) < r\}$. $r$ is a hyper-parameter that is set to 16 in this paper. Meanwhile, $score_a$ represents the score of the detected point $a$.

\section{Experiment}
\subsection{Dataset}
A subset of VisDrone \cite{zhuvisdrone2018} and TinyPerson\cite{Yu2020ScaleMF} are chosen to evaluate of CoarsePoint's performance.

\noindent\textbf{TinyPerson.} TinyPerson is a tiny object detection dataset collected from high-quality videos and pictures. It contains 72,651 annotated low-resolution human objects in 1,610 images, and most annotated objects in TinyPerson are smaller than $32\times32$ pixels. In the process of training and inference, the sub-image cut from the original image is used as input. In order to avoid incomplete objects caused by cutting, there is a constant overlap between adjacent sub-images. Then, the NMS strategy is used to fuse the same results of an image.

\noindent\textbf{Visdrone-Person.} As a large-scale dataset captured by UAV, Visdrone contains four tracks: (1) Image object detection, (2) Video object detection, (3) Single object tracking, and (4) Multi-object tracking. The experiments in this paper are conducted on the image object detection track of Visdrone. Considering the application scenarios of localization, we mainly focus on the object with a relatively small size. Therefore, the images containing human beings are used to construct a new human detection dataset named Visdrone-Person, and only the labels of pedestrians and persons are applied in this dataset. Visdrone-Person contains 10,209 images, with 6,471 images used for training, 548 for validation, and 3,190 for testing. The same cutting strategy used by TinyPerson is employed to get the sub-images with the appropriate size.

\subsection{Point Annotation Initialization} \label{5.2}
We obtained a point $(x, y)$ by Uniform Distribution or with the Rectified Gaussian Distribution within a bounding box $(x_c, y_c, w, h)$ to generate point annotations.

\noindent\textbf{Uniform Distribution.} Supposing manual annotating workers do not have any preference when labeling objects, the position of an annotated point follows the uniform distribution.

\noindent\textbf{Rectified Gaussian Distribution (RG).} Mostly, manual annotation does have a preference(\eg, mainly on the head or body). According to the large number theorem, it is reasonable to assume that the annotation labeled by a man is like a Gaussian distribution. However, Gaussian distribution has two problems when modelling the position distribution of labeled points. One is how to determine the variance of a Gaussian distribution. The other is that the range of the labeled points is bounded ($[- 0.5, 0.5]$), while the Gaussian is unbounded. Therefore, the RG distribution ($RG(x', y';\mu,\sigma)$) is adopted to handle these problems.

In this paper, the Uniform and $RG(x', y'; 0, 1/4)$ distributions are chosen to generate point annotations for most experiments. $RG(x', y';-0.26, 1/8)$, $RG(x', y';-0.28, 1/6)$, $RG(x', y';-0.31, 1/5)$, and $RG(x', y';-0.38, 1/4)$, whose mean are $-0.25$, are used for ablation study.

\subsection{RG Distribution}
First, as for the probability density function $G(x, y;\mu,\sigma)$ of Gaussian distribution, $\mu=0$. Following $G(x, y;\mu,\sigma)$ while sampling, the physical meaning of the sample falling outside of the range $[-0.5, 0.5]$ indicates that there is a certain probability that the position annotated by a man will be labeled outside the bounding box of objects that is often very small. According to the two sigma-rule and the three-sigma rule, it is generally believed that the sampling from Gaussian distribution falls on $[\mu-2\sigma, \mu+2\sigma]=[-2\sigma, 2\sigma]$ or $[\mu-3\sigma, \mu+3\sigma]=[-3\sigma, 3\sigma]$ is a low probability event, the probabilities of which are 4.56\% and 0.26\%  respectively. Consequently, we assume that the probability $P\{|x|>0.5\}=P\{|x|>\sigma*r\}$, in which $r = 2\ or\ 3$ depending on the labeling accuracy of the worker men. The variance has been determined as $\sigma=\frac{1}{2r}=1/4, 1/6$.

Second, think back to the process of human labeling. When humans label a point that falls outside an object, it is easy to find out. In this case, the man who labels the wrong annotated point should modify and relabel it until the point falls on the object. To lessen the problem, we assume that the distribution of the labeled point's position still conforms to the same Gaussian distribution. The probability of the annotated point fallings outside the object(bounding box is adopted to approximate object) of the first labeling $t$ can be obtained as $t=1-\int_{-0.5}^{0.5}\int_{-0.5}^{0.5}{G(x, y;\mu,\sigma)dxdy}$. We name the probability density function of the final annotated points' distribution as RG distribution($RG(x, y;\mu,\sigma)$). For $|x| \le 0.5\ and\ |y| \le 0.5,$ , the RG distribution can be deduced as follow: 

\begin{equation}
\label{eq6}
\begin{aligned}
RG(x, y;\mu, \sigma)&=G(x, y; \mu, \sigma) + t*G(x, y; \mu, \sigma) +\\
& \ \ ... + t^n*G(x, y; \mu, \sigma) + ...\\
& = G(x, y; \mu, \sigma) * \mathop{lim}\limits_{n->\infty}\frac{(1-t^n)}{1-t} \\
& = G(x, y; \mu, \sigma) / (1 - t) \\
& =\frac{G(x, y; \mu, \sigma)}{\int_{-0.5}^{0.5}\int_{-0.5}^{0.5}G(x, y; \mu, \sigma)dxdy},
\end{aligned}
\end{equation}
Otherwise, for $|x| > 0.5\ or\ |y| > 0.5,$ $RG(x, y;\mu, \sigma)=0$. Therefore, we adopt the RG distribution bounded on $[-0.5, 0.5]$ when generating point annotation from each object. (For $RG(x, y;\mu, \sigma)$, $\mu$ and ${\sigma}^2$ are not means and variance anymore.)

\subsection{Experimental Details}
\noindent\textbf{Implementation.} As a baseline model, RepPoints \cite{yang2019reppoints} is suitable for coarse point supervised object localization. This comprehensive framework consists of one estimator and one locator. 
To ensure that the proportions of positive and negative samples in training are appropriate, we regard some points around the labeled points as positive samples during label assigning in training. The network outputs the classification results and the position offset of each point. In practice, simply treat the input point as a pseudo box with a fixed length and width centered on the input point. Then, for self-refinement, the pseudo box is regarded as the supervision to train an estimator, after which the refined pseudo bounding box is fed to the locator. Finally, predicted bounding boxes produced by the locator are converted to the localization points in the evaluation.

\noindent\textbf{Setting.} 
ResNet-50 is used as a backbone network. The training epoch is set to 12. The initial learning rate is 0.01, and it decreases with the factor 0.1 in the 8$^{th}$ and 11$^{th}$ epoch. We choose Faster RCNN with FPN and Sparse RCNN\cite{sun2020sparse} for comparison.

\noindent\textbf{Metric.} We adopt the Average Precision (AP) as the metric. According to the bounding-box size, the scale interval is divided into tiny[2,20], small[20,32], normal[32,+$\infty$], and all[2,+$\infty$]. In point evaluation, the point-box distance, instead of IOU, is used as the matching criterion. 
Specifically, the threshold of point-box distance $\tau$ is set as 1. It means that as long as the predicted point falls within an unmatched ground truth, the match between the predicted point and the ground truth is successful.

\subsection{Experimental Analysis}

\begin{figure*}
\begin{center}
  \includegraphics[width=0.7\linewidth]{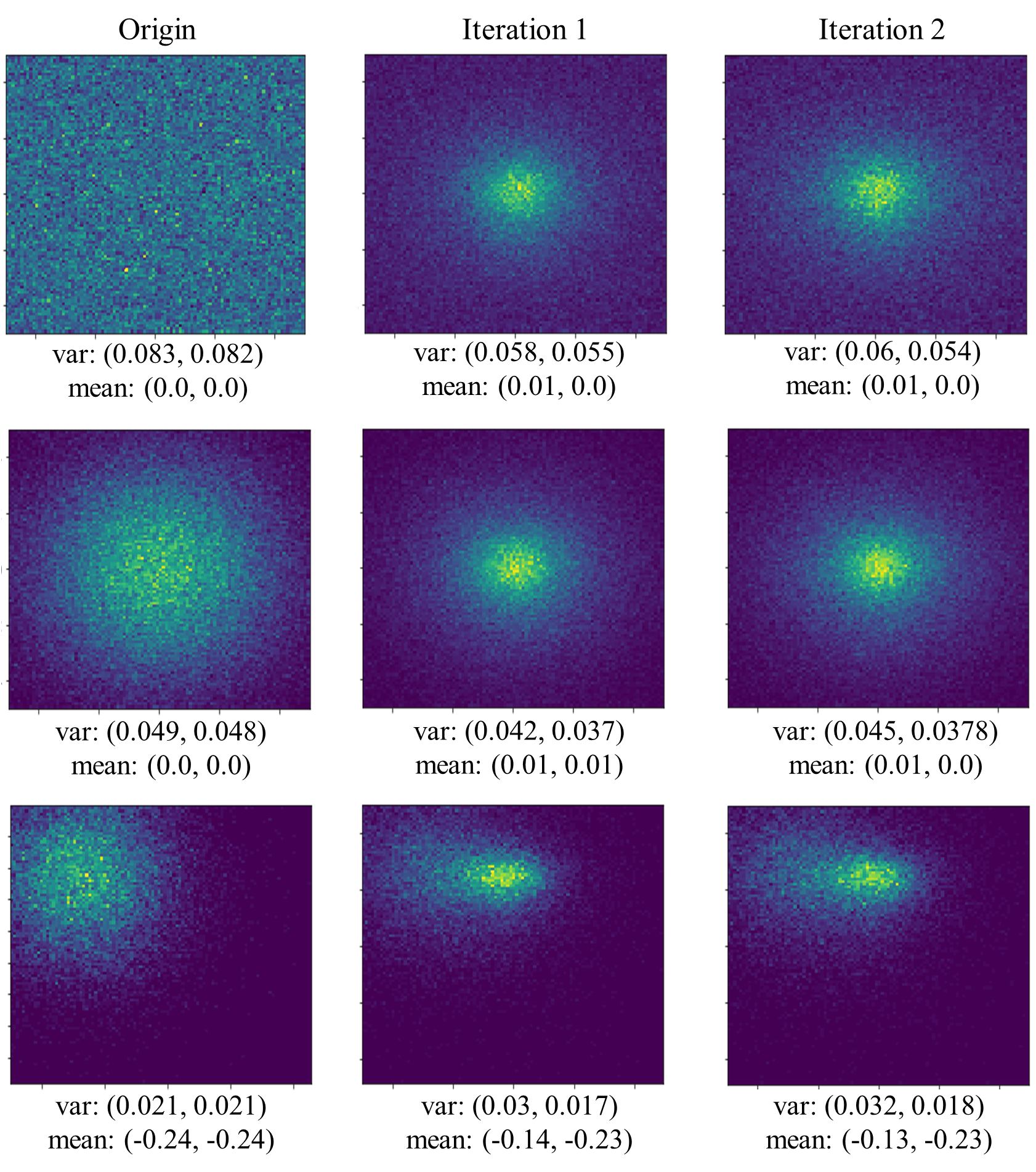}
\end{center}
\vspace{-4mm}
  \caption{
  Heatmaps of the point distribution in an object's bounding box. These three rows of heat maps indicate that the initial coarse point annotation obeys Uniform, RG(0, 1/4), and RG(-0.28, 1/6) distribution. The relevant performance data of these distributions that the initial coarse points obey can be seen in Tab.~\ref{Fig:different distribution}.
  }
\label{Fig:different distribution}
\end{figure*}

\setlength{\tabcolsep}{8pt}
\begin{table}
\normalsize
\begin{center}
\begin{tabular}{l|c|c|c}
\hline
\Xhline{1.2pt}
\multirow{2}{*}{Initial Distribution} & \multirow{2}{*}{std}& \multicolumn{2}{c}{\textbf{$AP_{1.0}^{all}$}}\\
\cline{3-4}
& & w/o. refined & w. refined \\
\hline
\Xhline{1.2pt}
\rowcolor{mygray}$Uniform$ & 0.289 & 59.7 & 68.5\\
$RG(0, 1/4)$ & 0.250 & 67.7 & 70.8\\
\rowcolor{mygray}$RG(0, 1/6)$ & 0.167 & 71.7 & 72.0\\
$RG(-0.38, 1/4)$ & 0.174 & 66.6 & 68.3\\
\hline
\Xhline{1.2pt}
\end{tabular}
\end{center}
\vspace{-2mm}
\caption{Comparisons of \textbf{$AP_{1.0}^{all}$} on VisDrone-Person under different initial distribution settings. The results predicted by RepPoints are obtained after two self-refinement iterations.}
\label{distribution}
\end{table}

\begin{table}
\begin{center}
\setlength{\tabcolsep}{8pt}{
\normalsize
\begin{tabular}{l|c|c|c}
\hline
\Xhline{1.2pt}
\multirow{2}{*}{Point Distribution} & \multirow{2}{*}{Std}& \multicolumn{2}{c}{\textbf{$AP_{1.0}^{all}$}}\\
\cline{3-4}
& & w/o. refined & w. refined \\
\hline
\Xhline{1.2pt}
\rowcolor{mygray}$RG(-0.38, 1/4)$ & 0.289 &  66.6 & 68.3\\
$RG(-0.31, 1/5)$ & 0.157 & 62.3 & 70.1\\
\rowcolor{mygray}$RG(-0.28, 1/6)$ & 0.142 & 68.0 & 69.7\\
$RG(-0.26, 1/8)$ & 0.117 & 70.2 & 72.0\\
\hline
\Xhline{1.2pt}
\end{tabular}}
\end{center}
\vspace{-2mm}
\caption{This series of experiments prove that the smaller the standard deviation of the RG distribution, the better the performance with the same initial mean of -0.25. We adopt RepPoints as the locator and the estimator with two self-refinement iterations on VisDrone-Person.}
\end{table}

\begin{figure*}
\begin{center}
    \begin{tabular}{ccc}
    \includegraphics[width=16cm]{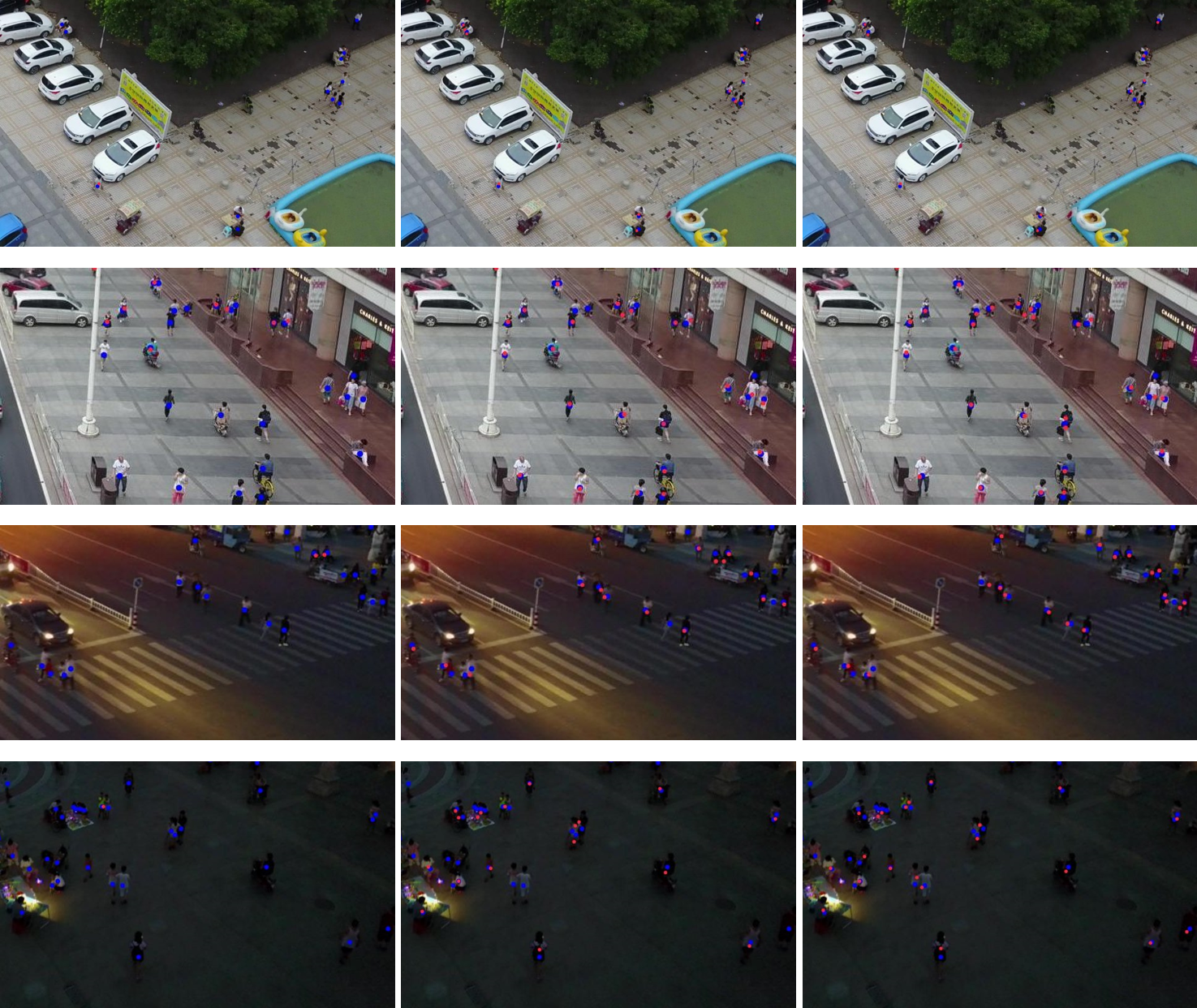}
    \end{tabular}
    \vspace{-2mm}
   \caption{The visualization of the estimator $E$'s results during different iterations on VisDrone-Person when the Recall was set as 0.5. We obtain the first column results by training $E$ without the self-refinement strategy. The results of the second column are obtained by $E$ after one iteration. Also, the results of the last column are obtained by $E$ after two iterations. Red points are the detected results, and blue points are the center of the ground-truth bounding boxes.}
\label{fig:estimate result}

\end{center}
\end{figure*}

\begin{figure*}
\begin{center}
    \begin{tabular}{ccc}
    \includegraphics[width=17cm]{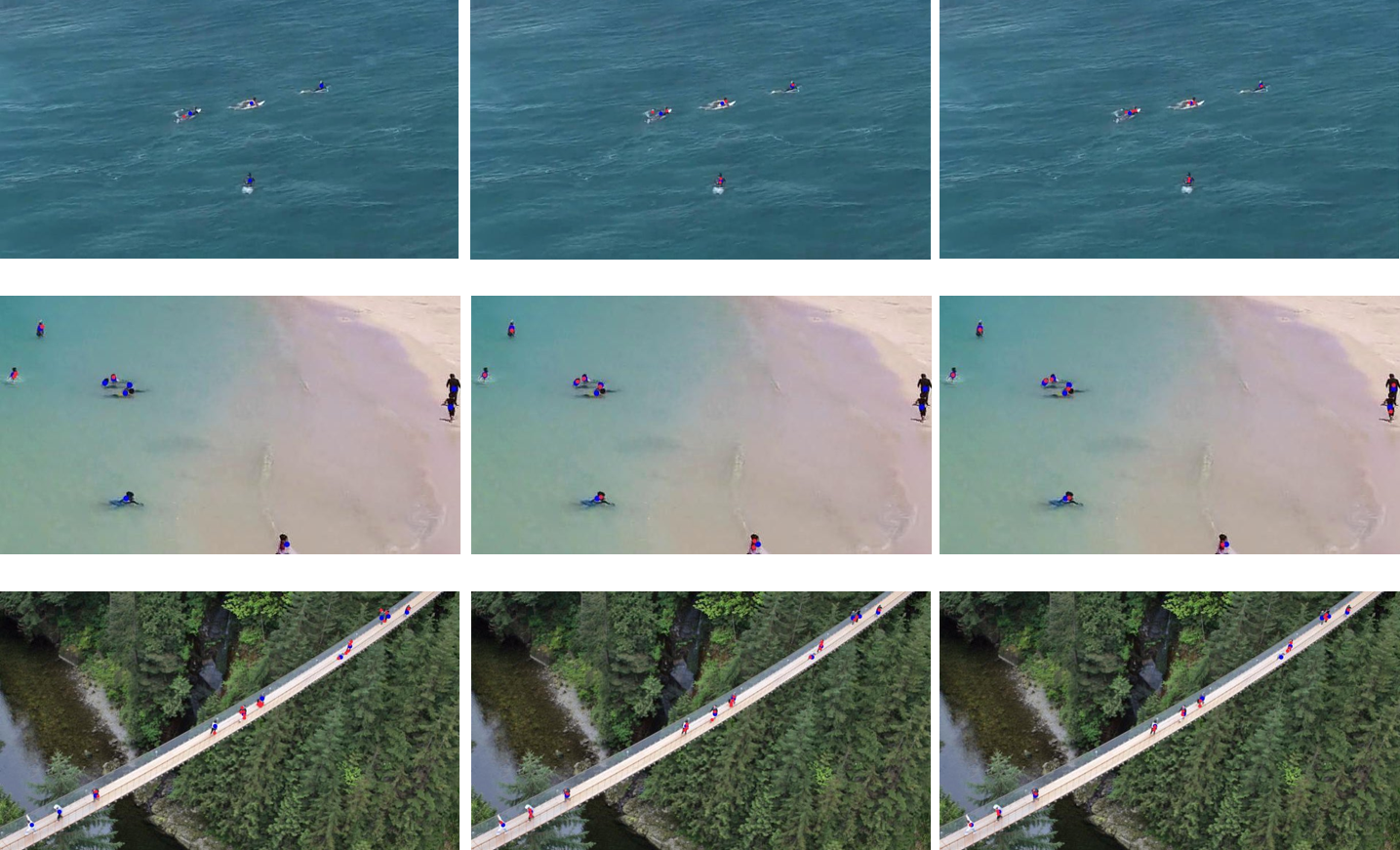}
    \end{tabular}
    \vspace{-2mm}
  \caption{The visualization of the estimator $E$'s results during different iterations on TinyPerson when the Recall was set as 0.5. We obtain the first column results by training $E$ without the self-refinement strategy. The results of the second column are obtained by $E$ after one iteration. Also, the results of the last column are obtained by $E$ after two iterations. Red points are the detected results, and blue points are the center of the ground-truth bounding boxes.}
\label{fig:estimate result}

\end{center}
\end{figure*}

\noindent\textbf{Decline of Model Confusion.} As mentioned in Sec.~\ref{sec3}, the uncertainty of annotation semantics limits the localization performance during the self-refinement. To quantitatively analyze this uncertainty, we collect the refined points in each iteration and calculate their relative positions $(x', y')$ within the object's bounding box. The distribution is defined as Eq \eqref{Eq:point's distribution}:
\begin{equation}
\begin{aligned}
&P(x', y'; A^k)\\
&= \frac{number\ of\ instances\ annotated\ on\ (x', y')}{\ number\ of\ instances},
\end{aligned}
\label{Eq:point's distribution}
\end{equation}
in which $|x'|\le 0.5$ and $|y'|\le 0.5$.

As shown in Fig.~\ref{Fig:different distribution}, after the self-refinement, the variance of the distribution decreases, which results in less annotation uncertainty and alleviates the model confusion. In the bottom line of Fig.~\ref{Fig:different distribution}, it is noticeable that with a small distribution variance and the initial annotation concentrated on the weaker semantic region, such as the top-left parts of persons, the proposed approach can refine the annotation to the neighboring region with strong semantic information, such as the head of a person.

\noindent\textbf{Reduction of False Positives.} Fig.~\ref{fig:estimate result} shows that the self-refinement significantly reduces the number of false positives, leading to the improvement of the network's discriminative ability.

\noindent\textbf{Analysis of Upper Bound.} As shown in Tab.~\ref{tab:error analysis},  we perform experiments based on VisDrone-Person with RepPoints as both the estimator and the locator. The real box means that precision bounding box annotations are used as the supervision information during training, and that the center point of the box was adopted for point evaluation. Box center represents using ($x_c$, $y_c$) as coarse point annotation during training, and it is the center of the original bounding box annotations. For box head, box foot, and box corner, we adopt ($x_{c}$, $y_{c}-h/4$), ($x_{c}$, $y_{c}+h/4$), and ($x_{c}-w/4$, $y_{c}-h/4$) respectively as point annotations. 

\setlength{\tabcolsep}{8pt}
\begin{table}
\begin{center}
\normalsize
\begin{tabular}{c|c|c|c|c}
\hline
\Xhline{1.2pt}
iteration & $AP^{all}_{1.0}$ & $AP^{tiny}_{1.0}$ & $AP^{small}_{1.0}$ & $AP^{normal}_{1.0}$  \\
\hline
\Xhline{1.2pt}
\rowcolor{mygray}0 & 60.2 & 61.9 & 46.9 & 19.1 \\
1 & 68.2 & 61.3 & 72.3 & 55.2\\
\rowcolor{mygray}2 & 68.9 & 60.7 & 71.9 & 62.9\\
\hline
\Xhline{1.2pt}
\end{tabular}
\end{center}
\vspace{-2mm}
\caption{The performance in VisDrone-Person with coarse point annotation following Uniform distribution, and RepPoints as the estimator and the locator.
}
\end{table}

\setlength{\tabcolsep}{7pt}
\begin{table}
\begin{center}
\normalsize
\begin{tabular}{l|c|c|c|c|c}
\hline
\Xhline{1.2pt}
Est & Loc & $AP^{all}_{1.0}$ & $AP^{tiny}_{1.0}$ & $AP^{small}_{1.0}$ & $AP^{normal}_{1.0}$\\
\hline
\Xhline{1.2pt}
\rowcolor{mygray}\ \ -  & RP & 59.7& 62.5& 44.5 & 14.0\\
RP & RP & 68.5& 60.6& 72.4 & 56.9\\
\rowcolor{mygray}\ \ -  & SR & 54.3 & 48.2 & 33.8 & 19.4\\
RP & SR & 67.5 & 57.1 & 59.7 & 43.6\\
\hline
\Xhline{1.2pt}
\end{tabular}
\end{center}
\vspace{-2mm}
\caption{Comparisons of APs with different locators on VisDrone-Person w self-refinement \textit{vs. }w/o self-refinement. The estimator is simplified as Est, the locator is simplified as Loc, RP represents RepPoints, SR represents Sparse RCNN, and - represents no self-refinement policy.}
\label{locators}
\end{table}

\setlength{\tabcolsep}{7pt}
\begin{table}
\begin{center}
\normalsize
\begin{tabular}{l|c|c|c|c|c}
\hline
\Xhline{1.2pt}
Est & Loc & $AP^{all}_{1.0}$ & $AP^{tiny}_{1.0}$ & $AP^{small}_{1.0}$ & $AP^{normal}_{1.0}$\\
\hline
\Xhline{1.2pt}
\rowcolor{mygray}\ \  -  & FR & 45.8 & 49.3 & 15.3 & 10.2 \\
RP & FR & 63.4 & 63.2 & 41.5 & 19.9\\
\rowcolor{mygray}FR & FR & 66.1 & 66.3 & 47.6 & 18.9\\
\hline
\Xhline{1.2pt}
\end{tabular}
\end{center}
\vspace{-2mm}
\caption{Comparisons of APs with different estimators on TinyPerson. FR represents Faster RCNN.}
\label{estimators}
\end{table}

\subsection{Ablation Study}
\noindent\textbf{Locators.} In Tab.~\ref{locators}, self-refinement improves $AP^{all}_{1.0}$ and $AP^{normal}_{1.0}$ by 8.82 and 42.9 points respectively under RP+RP conditions. Meanwhile, $AP^{all}_{1.0}$ and $AP^{normal}_{1.0}$ are increased by 13.2 and 24.2 points respectively over RP+SR.

\noindent\textbf{Estimators.} It is shown in Tab.~\ref{estimators} that even with different estimators, self-refinement improves $AP^{all}_{1.0}$ by 20.27 points and $AP^{small}_{1.0}$ by 32.28 points with FR+FR. Our self-refinement algorithm is compatible with different frameworks. When different detectors are used as estimators or locators, the performance under various scales has almost been improved.

\noindent\textbf{Dataset.} As shown in Tab.~\ref{locators} and Tab.~\ref{estimators}, the performance of the self-refinement showed 8.2 and 20.3 points increase on VisDrone-Person and TinyPerson, respectively.

\noindent\textbf{Number of Iteration.} The results in Tab.~\ref{iter_box} reveal that the performance significantly improves as the number of iterations increases. Then, $AP_{1.0}^{all}$ gains 30.6, 16.87 and 7.33 points in the initial pseudo box size setting with 8*8, 16*16 and 32*32 pixels, respectively. Alternatively stated, the greater the number of iterations is, the slower the performance grows. 

\noindent\textbf{Pseudo Box Size.} Tab.~\ref{iter_box} also depicts the effect of different initial pseudo box sizes. Increasing the pseudo box size from 8*8 pixels to 32*32 pixels shows that a larger pseudo box size achieves a better performance. However, self-refinement with more iterations drives a performance boost.

\noindent\textbf{Annotation Initialization.} In Tab.~\ref{distribution}, the results show that the initial distribution of generated points affects performance. Manual annotation is more consistent with the RG distribution.
The improvement brought by self-refinement is greater when encountering coarser annotation. In addition, the performance becomes relatively stable and insensitive to the initial annotation distribution through self-refinement.

\setlength{\tabcolsep}{18.5pt}
\begin{table}
\begin{center}
\normalsize
\begin{tabular}{c|c|c|c}
\hline
\Xhline{1.2pt}
\multirow{2}{*}{Iteration} &  \multicolumn{3}{c}{\textbf{$AP_{1.0}^{all}$}}\\
\cline{2-4}
 & 8*8 & 16*16 & 32*32 \\
\hline  
\Xhline{1.2pt}
\rowcolor{mygray}0 & 21.4 & 45.82 & 58.92 \\
1 & 52.0 & 62.69 & 66.25 \\
\rowcolor{mygray}2 & 60.0 & 66.09 & 66.08 \\
3 & 62.8 & -     & -     \\
\hline
\Xhline{1.2pt}
\end{tabular}
\end{center}
\vspace{-2mm}
\caption{Results on TinyPerson after different iterations and in the different initial pseudo box sizes with Faster RCNN as the estimator and Faster RCNN as the locator.}
\label{iter_box}
\end{table}

\setlength{\tabcolsep}{6.5pt}
\begin{table}
\begin{center}
\normalsize
\begin{tabular}{l|c|c|c|c}
\hline
\Xhline{1.2pt}
Initial Point & $AP^{all}_{1.0}$ & $AP^{tiny}_{1.0}$ & $AP^{small}_{1.0}$ & $AP^{normal}_{1.0}$  \\
\hline
\Xhline{1.2pt}
\rowcolor{mygray}real box & 76.2 & 65.0 & 80.7 & 82.6\\
box center& 74.1 & 63.5 & 79.1 & 76.8\\
\rowcolor{mygray}box head & 71.2 & 60.2 & 75.4 & 72.5 \\
box foot & 72.7 & 61.0 & 77.7 & 74.0\\
\rowcolor{mygray}box corner & 68.6 & 63.6 & 68.2 & 38.2 \\
\hline
\Xhline{1.2pt}
\end{tabular}
\end{center}
\vspace{-2mm}
\caption{Upper bound analysis. This experiment shows the performance of different kinds of semantic points generated from the bounding box on VisDrone-Person.} 
\label{tab:error analysis}
\end{table}

\subsection{Annotation Efficiency}
To quantitatively compare annotation efficiencies mentioned above, we randomly select some images in different scenarios from TinyPerson and VisDrone-Person with 1021 objects, and nine persons are chosen for conducting a manual annotation test.
In order to avoid the influence of the annotation order, the testers are randomly divided into three groups. Any group is selected to annotate these images following the order of coarse point, precision point, bounding box. Another group follows the order of precision point, coarse point, bounding box. The third group follows the order of bounding box, precision point, coarse point. Finally, we calculate the single object's average annotation time of different annotations, shown in Tab.~\ref{annotation efficiecy}.

The annotation of tight bounding boxes is time-consuming, especially for finding small objects in large-scale images. The annotation of the coarse point is quite efficient, and it just needs to click on the object. Under such annotation, our proposed self-refinement algorithm can still achieve excellent localization performance while saving much annotation time up to 80$\%$.

\setlength{\tabcolsep}{16.5pt}
\begin{table}
\begin{center}
\normalsize
\begin{tabular}{l|c|c}
\hline
\Xhline{1.2pt}
Annotation Type & Average Time & $AP_{1.0}^{all}$ \\
\hline
\Xhline{1.2pt}
\rowcolor{mygray}bounding box    & 9.4s & 76.2 \\
precision point & 5.3s & 71.2 \\
\rowcolor{mygray}coarse point    & 1.8s & 70.8 \\
\hline
\Xhline{1.2pt}
\end{tabular}
\end{center}
\vspace{-2mm}
\caption{Annotation efficiency. We counted the average time and evaluated the final performance of the same group of people using different annotation types to label images from VisDrone-Person. The performance of coarse points is obtained with reference to the RG(0,1/4) distribution. It only lost 5, 0.4 points of $AP_{1.0}^{all}$ while saving the annotation cost up to 80\% and 66\% in comparison with bounding-box and precision point correspondingly.} 
\label{annotation efficiecy}
\end{table}

\section{Conclusion}
Object localization is important in the computer vision community. In this paper, we propose CoarsePoint, a new vision task for object localization with coarse point supervision, setting the first solid baseline for the community. Furthermore, we propose the self-refinement approach to promote coarse points iteratively, implementing the statistical stability of supervision signals and improving the localization model in a self-paced fashion. CoarsePoint achieves comparable experimental results with precise bounding-box annotation-based methods and saves up to 80$\%$ annotation cost.



%


\section{ACKNOWLEDGMENT}
This work was supported by the National Natural Science Foundation of China (NSFC) under Grant No. 61836012 and 61771447, and the Strategic Priority Research Program of the Chinese Academy of Sciences under Grant No. XDA27000000.

\ifCLASSOPTIONcaptionsoff
\newpage
\fi

\footnotesize
\bibliographystyle{IEEEtran}
\bibliography{IEEEabrv,paper}



\vspace{-10mm}

\begin{IEEEbiography}[{\includegraphics[width=1in,height=1.25in,clip,keepaspectratio]{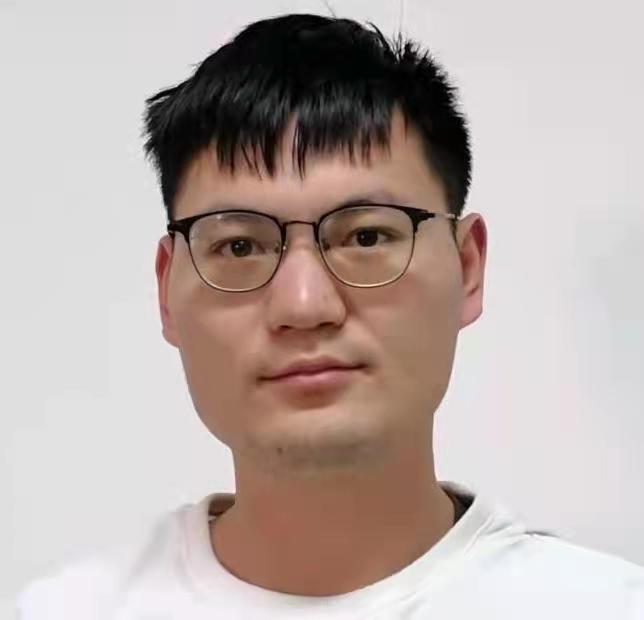}}]{Xuehui Yu}
received the B.E. degree in software engineering from Tianjin University, China, in 2017. He is currently pursuing the Ph.D. degree in signal and information processing with University of Chinese Academy of Sciences. His research interests include machine learning and computer vision.
\end{IEEEbiography}

\vspace{-10mm}

\begin{IEEEbiography}[{\includegraphics[width=1in,height=1.25in,clip,keepaspectratio]{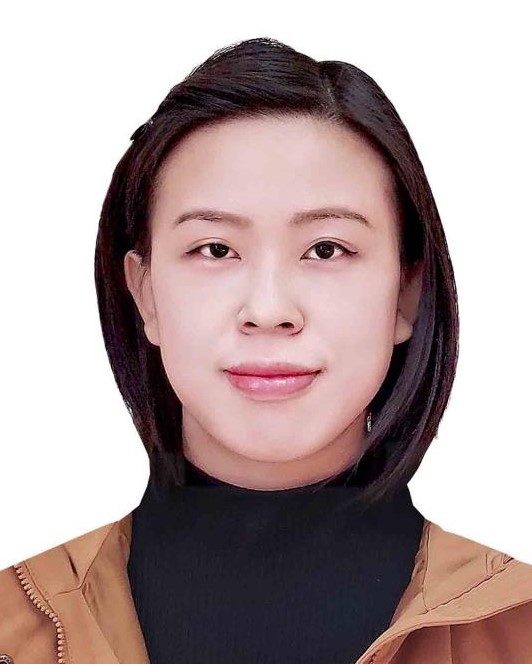}}]{Di Wu}
received the B.E. degree in software engineering from Tianjin University, China, in 2019. She is currently pursuing the M.S. degree in electronic and communication engineering with University of Chinese Academy of Sciences. Her research interests include machine learning and computer vision.
\end{IEEEbiography}



\vspace{-10mm}

\begin{IEEEbiography}[{\includegraphics[width=1in,height=1.25in,clip,keepaspectratio]{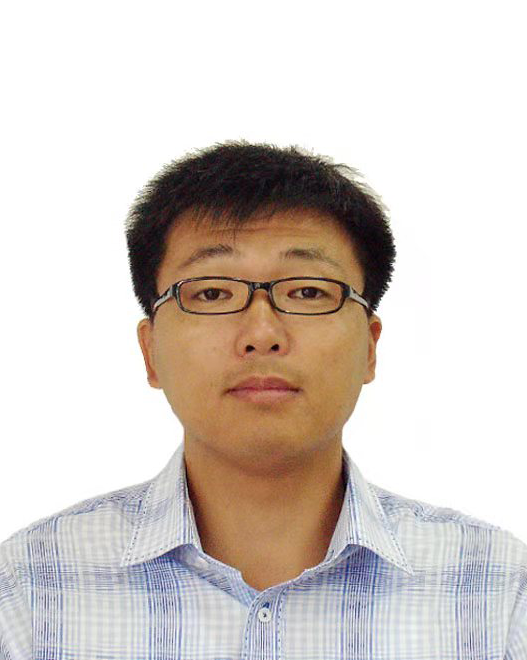}}]{Qixiang Ye}
received the B.S. and M.S. degrees from Harbin Institute of Technology, China, in 1999 and 2001, respectively, and the Ph.D. degree from the Institute of Computing Technology, Chinese Academy of Sciences in 2006. He has been a professor with the University of Chinese Academy of Sciences since 2009, and was a visiting assistant professor with the Institute of Advanced Computer Studies (UMIACS), University of Maryland, College Park until 2013. His research interests include image processing, visual object detection and machine learning. He has published more than 100 papers in refereed conferences and journals including IEEE CVPR, ICCV, ECCV and PAMI. He is on the editorial boards of IEEE Transactions on Intelligent Transportation System and IEEE Transactions on Circuit and System on Video Technology.
\end{IEEEbiography}

\vspace{-16mm}

\begin{IEEEbiography}[{\includegraphics[width=1in,height=1.25in,clip,keepaspectratio]{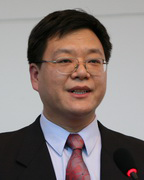}}]{Jianbin Jiao}
received the B.S., M.S., and
Ph.D. degrees in mechanical and electronic engineering
from Harbin Institute of Technology (HIT),
Harbin, China, in 1989, 1992, and 1995, respectively.
From 1997 to 2005, he was an Associate
Professor with HIT. Since 2006, he has been a
Professor with the School of Electronic, Electrical,
and Communication Engineering, University of the
Chinese Academy of Sciences, Beijing, China. His
current research interests include image processing,
pattern recognition, and intelligent surveillance.
\end{IEEEbiography}

\vspace{-16mm}

\begin{IEEEbiography}[{\includegraphics[width=1in,height=1.25in,clip,keepaspectratio]{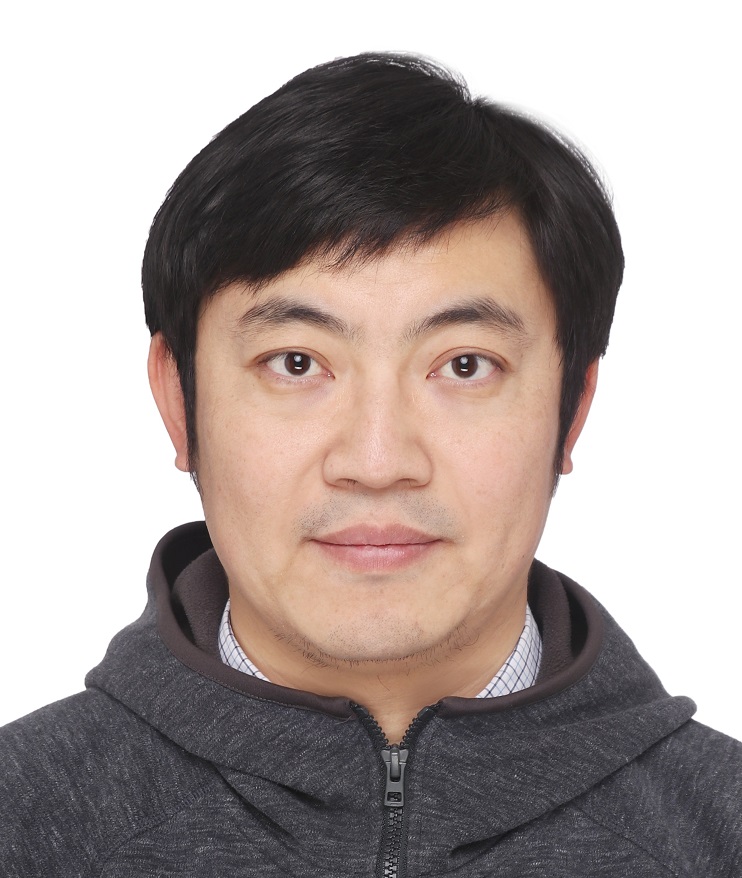}}]{Zhenjun Han}
received the B.S. degree in software engineering from Tianjin University, Tianjin, China, in 2006 and the M.S. and Ph.D. degrees from University of Chinese Academy of Sciences, Beijing, China, in 2009 and 2012, respectively. Since 2013, he has been an Associate Professor with the School of Electronic, Electrical, and Communication Engineering, University of Chinese Academy of Sciences. His research interests include object tracking and detection.
\end{IEEEbiography}

\end{document}